\documentclass[letterpaper, 10 pt, conference]{ieeeconf}  

\IEEEoverridecommandlockouts                      

\usepackage{epsfig} 
\usepackage{amsmath} 
\usepackage{amssymb}  
\usepackage{color}
\usepackage[linesnumbered,ruled]{algorithm2e}
\usepackage[normalem]{ulem} 
\makeatletter
\let\NAT@parse\undefined
\makeatother
\usepackage{hyperref}
\hypersetup{pdfstartview=FitH,
            colorlinks=true,
            linkcolor=red,
            anchorcolor=blue,
            citecolor=green
            }
\usepackage{cite}
\usepackage{booktabs}
\usepackage{multirow}
\usepackage[caption=false,font=footnotesize]{subfig}
\usepackage[dvipsnames]{xcolor}

\usepackage{soul}
\usepackage{balance}
\newcounter{RNum}
\renewcommand{\theRNum}{\arabic{RNum}}
\newcommand{\Remark}{\noindent\textit{\textbf{Remark}~\refstepcounter{RNum}\textbf{\theRNum}: }}
\newcommand{\NoOne}[1]{\textcolor{red}{#1}}
\newcommand{\NoTwo}[1]{\textcolor{green}{#1}}
\newcommand{\NoThree}[1]{\textcolor{blue}{#1}}
\soulregister\NoOne7
\soulregister\NoTwo7
\soulregister\NoThree7
\soulregister\Remark7

\title{\LARGE \bf
SGDViT: Saliency-Guided Dynamic Vision Transformer\\ for UAV Tracking
}

\author{Liangliang Yao$^{1}$, Changhong Fu$^{1,*}$, Sihang Li$^{1}$, Guangze Zheng$^{2}$, and Junjie Ye$^{1}$ 
\thanks{$^{*}$Corresponding author}%
\thanks{$^{1}$Liangliang Yao, Changhong Fu, Sihang Li, and Junjie Ye are with the School of Mechanical Engineering, Tongji University, Shanghai 201804, China.
        {\tt\footnotesize changhongfu@tongji.edu.cn}}%
\thanks{$^{2}$Guangze Zheng is with the Department of Computer Science, the University of Hong Kong, Hong Kong, China.}
}

\begin{document}

\maketitle
\thispagestyle{empty}
\pagestyle{empty}


\begin{abstract}

Vision-based object tracking has boosted extensive
autonomous applications for unmanned aerial vehicles (UAVs). However, the dynamic changes in flight maneuver and viewpoint encountered in UAV tracking pose significant difficulties, \textit{e.g.}, aspect ratio change, and scale variation. The conventional cross-correlation operation, while commonly used, has limitations in effectively capturing perceptual similarity and incorporates extraneous background information. To mitigate these limitations, this work presents a novel saliency-guided dynamic vision Transformer (SGDViT) for UAV tracking. The proposed method designs a new task-specific object saliency mining network to refine the cross-correlation operation and effectively discriminate foreground and background information. Additionally, a saliency adaptation embedding operation dynamically generates tokens based on initial saliency, thereby reducing the computational complexity of the Transformer architecture. Finally, a lightweight saliency filtering Transformer further refines saliency information and increases the focus on appearance information. The efficacy and robustness of the proposed approach have been thoroughly assessed through experiments on three widely-used UAV tracking benchmarks and real-world scenarios, with results demonstrating its superiority. The source code and demo videos are available at \url{https://github.com/vision4robotics/SGDViT}. 
\end{abstract}



\section{Introduction} \label{sec:intro}
Visual object tracking has gained increasing attention in various applications of aerial robotics, \textit{e.g.}, 3D localization~\cite{zhang2019eye}, aerial cinematography~\cite{bonatti2019IROS}, and target localization~\cite{Ye_2021_TIE}. UAV trackers' aim is to predict the object's location based on the template information provided in the initial frame. However, the frequent changes in flight maneuver and viewpoint encountered in UAV tracking present substantial challenges, \textit{e.g.}, aspect ratio change, and scale variation. These challenges can greatly decrease the perceptual similarity between the template and search region, resulting in a significant drop in the performance of the tracker. Despite some progress made in recent years, robust and effective UAV tracking remains a challenging problem.


Over the past few years, Siamese network-based trackers \cite{fu2022siamese,li2019siamrpn++,xu2020siamfc++,cao2021hift} have emerged as the dominant approach in UAV tracking. The method conceives tracking as a problem of template matching, \textit{i.e.}, adopting the initial frame's designated target as a reference for subsequent frames.
Cross-correlation is utilized to generate similarity maps by extracting features from both the template and the search region. The majority of Siamese trackers~\cite{bertinetto2016fully,li2019siamrpn++,xu2020siamfc++} apply these maps for the purposes of classification and regression. However, the cross-correlation operation is a linear matching process and brings about the infusion of background information when the object's appearance changes dramatically during UAV tracking, resulting in unstable performance.
To counteract this issue, a considerable number of Siamese trackers~\cite{zhu2018distractor,li2019target} have integrated feature refinement modules for enhancing the robustness of similarity feature extraction. Despite these attempts, the cross-correlation operation remains susceptible to similar background information, and its ability to extract perceptual similarity remains limited when dramatically appearance variations occur.


\begin{figure}[!t]	
	\centering
	\includegraphics[width=1\linewidth]{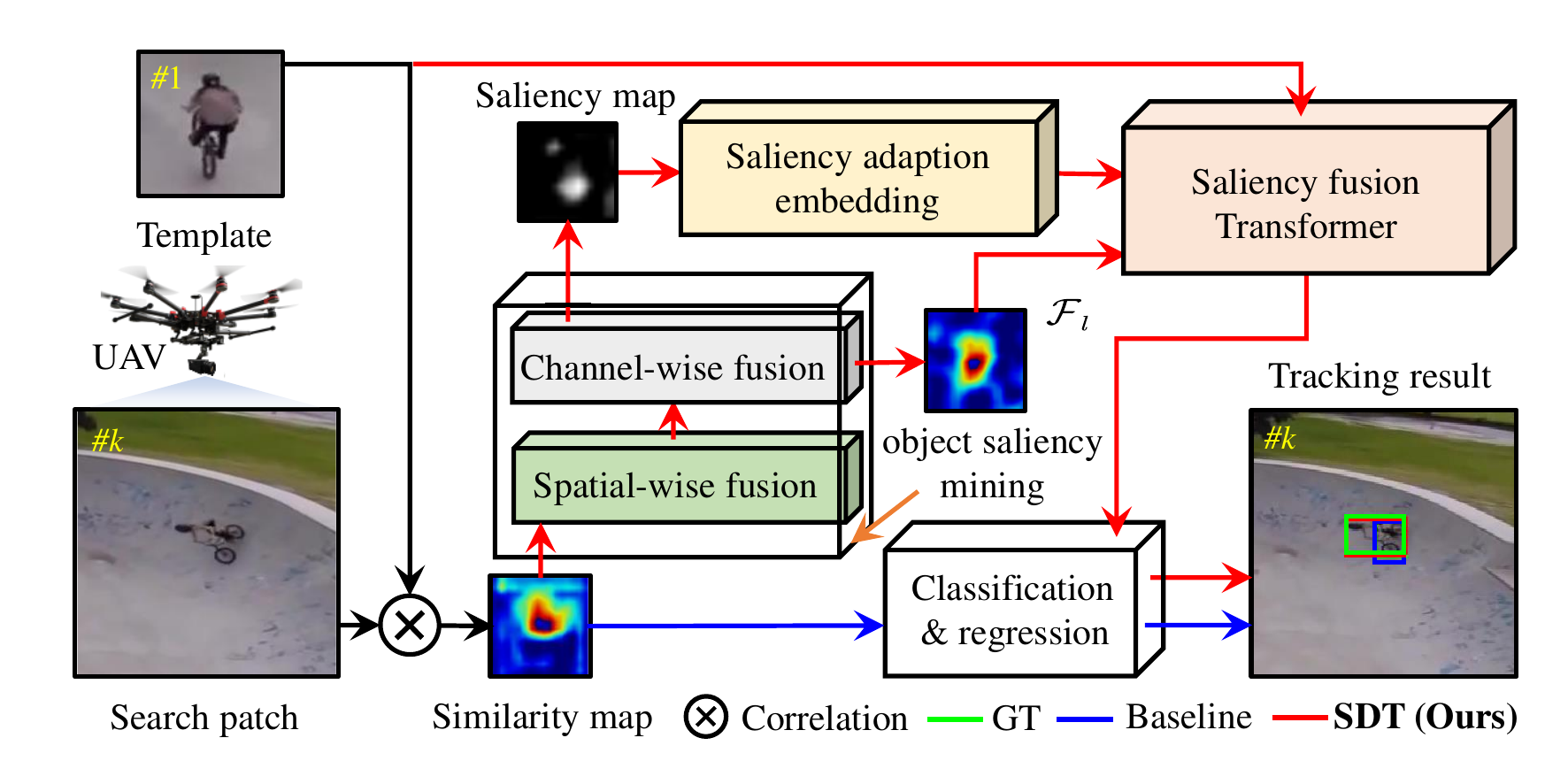}
	\setlength{\abovecaptionskip}{-14pt} 
	\caption
    {Overall comparison of tracking results between baseline (\textcolor[rgb]{0,0,1}{blue}) and the proposed saliency-guided dynamic vision Transformer (SGDViT) (\textcolor[rgb]{1,0,0}{red}). The $\mathcal{F}_l$ is the mined saliency feature. The \textcolor[rgb]{0,1,0}{green} box represents the ground-truth (GT) bounding box. (Image frames are from DTB70 \cite{li2017visual}.)
	}
	\label{fig:fig1}
\end{figure}

Recently, Transformer~\cite{vaswani2017nips} has shown tremendous potential in computer vision. In particular, the Transformer structure has exhibited superior performance in various scenarios of UAV tracking~\cite{cao2021hift,xing2022siamese}. To address the decrease in perceptual similarity, some researchers have attempted to replace the cross-correlation operation with Transformer~\cite{sun2020transtrack,wang2021transformer}. 
Nevertheless, the Transformer structure has quadratic computational complexity in terms of the number of tokens, which hinders its deployment on embedded UAV platforms with limited computational resources. Thus, reducing the computational complexity of the Transformer structure is a pressing problem for UAV tracking. In this work, a novel saliency-guided dynamic vision Transformer (SGDViT) for UAV tracking is proposed, as illustrated in Figure~\ref{fig:fig1}. 
The key contributions of this work are as follows:

\begin{itemize}
	
	\item A novel saliency-guided dynamic vision Transformer dubbed SGDViT is proposed to enhance the ability to extract perceptual similarity and mitigate the interference of background information, thereby addressing the issue of reduced perceptual similarity due to aspect ratio change and scale variation.
	
	\item An original object saliency mining module is designed to distinguish foreground and background by extracting salient information through an organic combination of spatial and channel fusion. 
	
	\item A new saliency adaption embedding operation is developed to accelerate the Transformer structure by incorporating tokens at various levels based on initial saliency information. On this base, a novel saliency fusion Transformer is implemented to refine the saliency information and minimize the loss of appearance information through the cross-attention mechanism.
	
	
	\item Comprehensive evaluations conducted on three authoritative UAV tracking benchmarks have demonstrated the robust performance of SGDViT in addressing appearance variations. Real-world tests further validate the efficiency and effectiveness of the presented approach.

\end{itemize}

\section{Related Works}
\subsection{Visual Object Tracking for UAV}
Previously, discriminative correlation filter (DCF) based trackers \cite{Ye_2021_TIE,li2020autotrack,fu2021correlation} have attracted great attention for their computational efficiency. However, their limited end-to-end training makes them ill-suited to handle complex situations, particularly aspect ratio change, and scale variation. Siamese network-based methods~\cite{bertinetto2016fully,li2018high,cao2021siamapn++,li2019siamrpn++} are the mainstream methods. As one of the pioneering works, SiamFC~\cite{bertinetto2016fully}, regarding the tracking task as the feature matching process between template and search regions, has a profound impact on the subsequent Siamese trackers. SiamAPN++~\cite{cao2021siamapn++} proposes an attention aggregation network (AAN) to improve feature representation ability. The network consists of two parts: Self-AAN and Cross-AAN, which can aggregate target features at adaptive scales. HiFT~\cite{cao2021hift} applies the Transformer to fuse multi-level features, thereby discovering a tracking-tailored feature space with strong discriminability. Recently, TransT~\cite{chen2021transformer} replaces the correlation-based network using the proposed feature fusion network inspired by Transformer. However, the decrease in perceptual similarity can lead to the introduction of unnecessary background information in the similarity map constructed through cross-correlation, resulting in unstable tracking performance. Despite the improved results obtained through the use of the Transformer, its quadratic computational complexity remains a significant obstacle for deployment on UAVs with limited computing resources.

\begin{figure*}[!t]	
	\centering
	\includegraphics[width=0.95\linewidth]{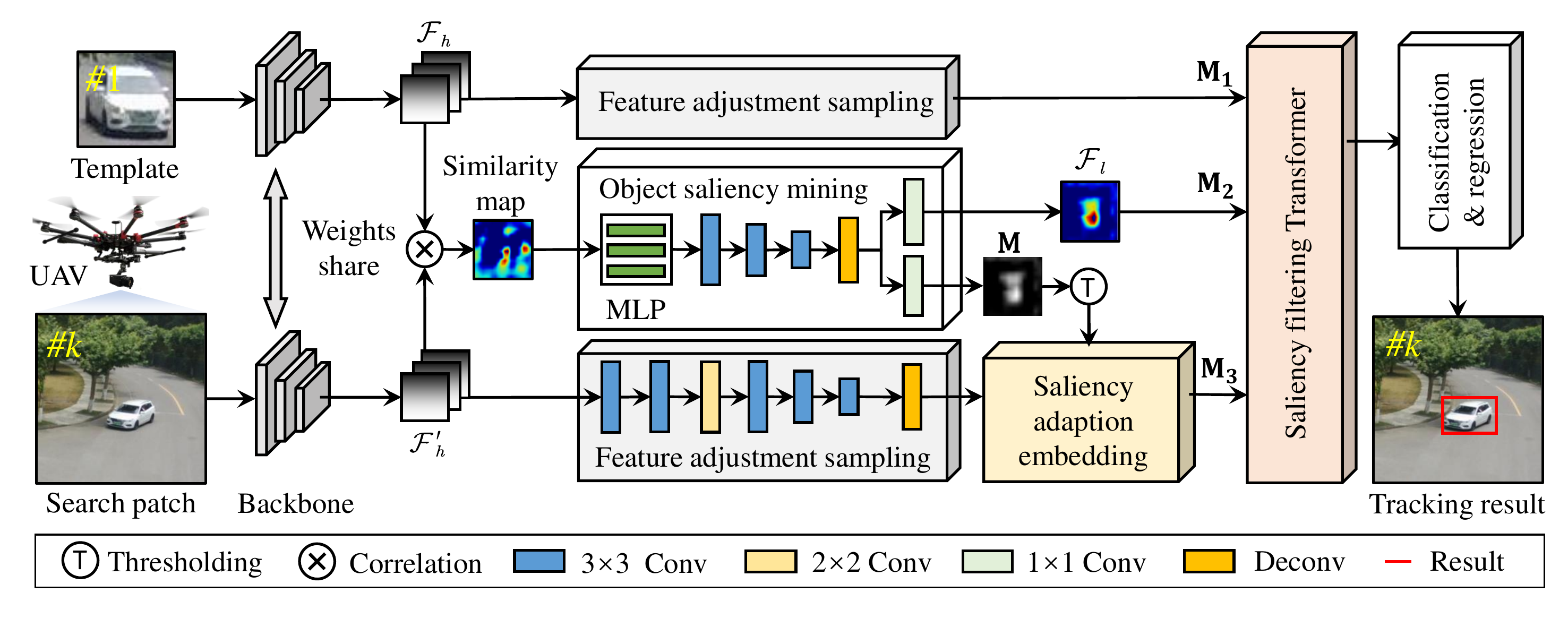}
	\setlength{\abovecaptionskip}{-1pt}

	\caption
	{
		Overview of the proposed SGDViT tracker.
		The parts from the left to right are the \emph{feature extraction network, object saliency mining network, feature adjustment sampling network, saliency filtering Transformer, and classification \& regression network}, respectively. For clarity, the feature maps of the template and search regions are respectively denoted by $\mathcal{F}_h$ and $\mathcal{F}^{'}_h$. The $\mathcal{F}_l$ is the mined saliency feature. (Image frames are from UAVTrack112 \cite{fu2021onboard}.)
	}
	\label{fig:main}
	
\end{figure*}

\subsection{Salient Object Detection}

Salient object detection, aiming at highlighting visually salient object regions in images, has attracted widespread concern in computer vision\cite{borji2019salient}. DSAR-CF \cite{feng2019dynamic} proposes a new dynamic saliency-aware regularized CF tracking method to guide the online updating of the regularization weight map.
SAOT \cite{zhou2021saliency} proposes a fine-grained saliency mining module to capture the local saliencies and designs a saliency-association modeling module to associate the captured saliencies together. In comparison with conventional salient object detection methods, the proposed object saliency mining network processes similarity maps rather than raw images. The proposed method demonstrates remarkable efficacy in extracting perceptual similarity and identifying object regions in a preliminary stage.

\subsection{Dynamic Token Generation}
The generation of dynamic tokens is a crucial aspect in enhancing the efficiency of the Transformer architecture for real-time applications. For this operation, a critical factor is the evaluation of the significance of each token. DynamicViT ~\cite{rao2021dynamicvit} devises a lightweight prediction module to estimate the importance score of each token and determine which tokens to be pruned dynamically. QuadTree Attention ~\cite{tang2022quadtree} builds token pyramids and computes attention according to the attention scores. This method skips irrelevant regions in the fine level if their corresponding coarse-level regions are not promising, thereby reducing the computational complexity from quadratic to linear. In contrast, the saliency adaption embedding operation this work proposes is specifically designed for UAV tracking missions. The significance of each token is determined through the mined saliency information, which can distinguish foreground and background.

\section{Proposed Method}
The proposed saliency-guided dynamic vision Transformer (SGDViT) is introduced in detail. As depicted in Fig.~\ref{fig:main}, the proposed framework can be partitioned into five parts, \textit{i.e.}, \emph{feature extraction network, object saliency mining network, feature adjustment sampling network, saliency filtering Transformer, and classification \& regression network}.

\subsection{Object Saliency Mining Network}
For clarity, the feature maps of the template and search regions are respectively denoted by $\mathcal{F}_h$ and $\mathcal{F}^{'}_h$. Specifically, the feature maps $\mathcal{F}_h$ and $\mathcal{F}^{'}_h$ are convoluted and reshaped to the similarity map $\mathbf{S}_{\rm 1} \in \mathbb{R}^{H \times W \times C}$, where $H, W, C$ represent the height, width, and channel of the feature map respectively. Subsequently, a spatial fusion layer and a channel fusion layer are used to further process the similarity map. The spatial fusion layer incorporates a multi-layer perceptron (MLP) network, while the channel fusion layer consists of a convolution-deconvolution network. The result of the channel fusion layer, denoted as $\mathbf{S}_{\rm 2}$, is given by: 
\begin{equation}\label{1}
    \begin{split}
        &\mathbf{S}_{\rm 1}=\mathcal{F}^{'}_h\star\mathcal{F}_h\quad,\\
        &\mathbf{S}_{\rm 2}=\mathrm{Deconv}(\mathrm{Conv}(\mathrm{MLP}(\mathbf{S}_{\rm 1}))) \quad,
    \end{split}
\end{equation}
where $\star$ represents the cross-correlation operator. The MLP network structure enables the communication of spatial information, while the convolutional and deconvolutional network structures facilitate the integration of information across channels. Afterward, the results $\mathbf{S}_{\rm 2}$ are passed through two different branches respectively to generate the saliency features $\mathcal{F}_l \in \mathbb{R}^{H \times W \times C}$ and the saliency map $\mathbf{M} \in \mathbb{R}^{H \times W \times 1}$ for the saliency filtering Transformer. The main reason for using two networks here is the different purposes. For the saliency map $\mathbf{M}$, our main idea is to compress the saliency information to the $1$ dimension, playing the role of cross-channel aggregation. For the saliency features $\mathcal{F}_l$, our main target is to refine the perceptual similarity further and perform cross-channel information interaction. 

\Remark The Multi-layer Perceptron (MLP) layers have a distinct advantage in capturing long dependencies and spatial relationships, whereas convolution layers excel in the extraction of information between local features and the fusion of channel information. The integration of these two layers enhances the ability to represent features. 


\subsection{Saliency Adaption Embedding Operation}

In preparation for dynamic token generation, a feature adjustment sampling network is designed to adjust the size of the feature, as illustrated in Fig.~\ref{fig:main}. The network comprises several convolution layers with varying receptive fields to effectively address UAV tracking challenges, specifically for aspect ratio change and scale variation.

\begin{figure}[!t]
	\centering
	\includegraphics[width=1\linewidth]{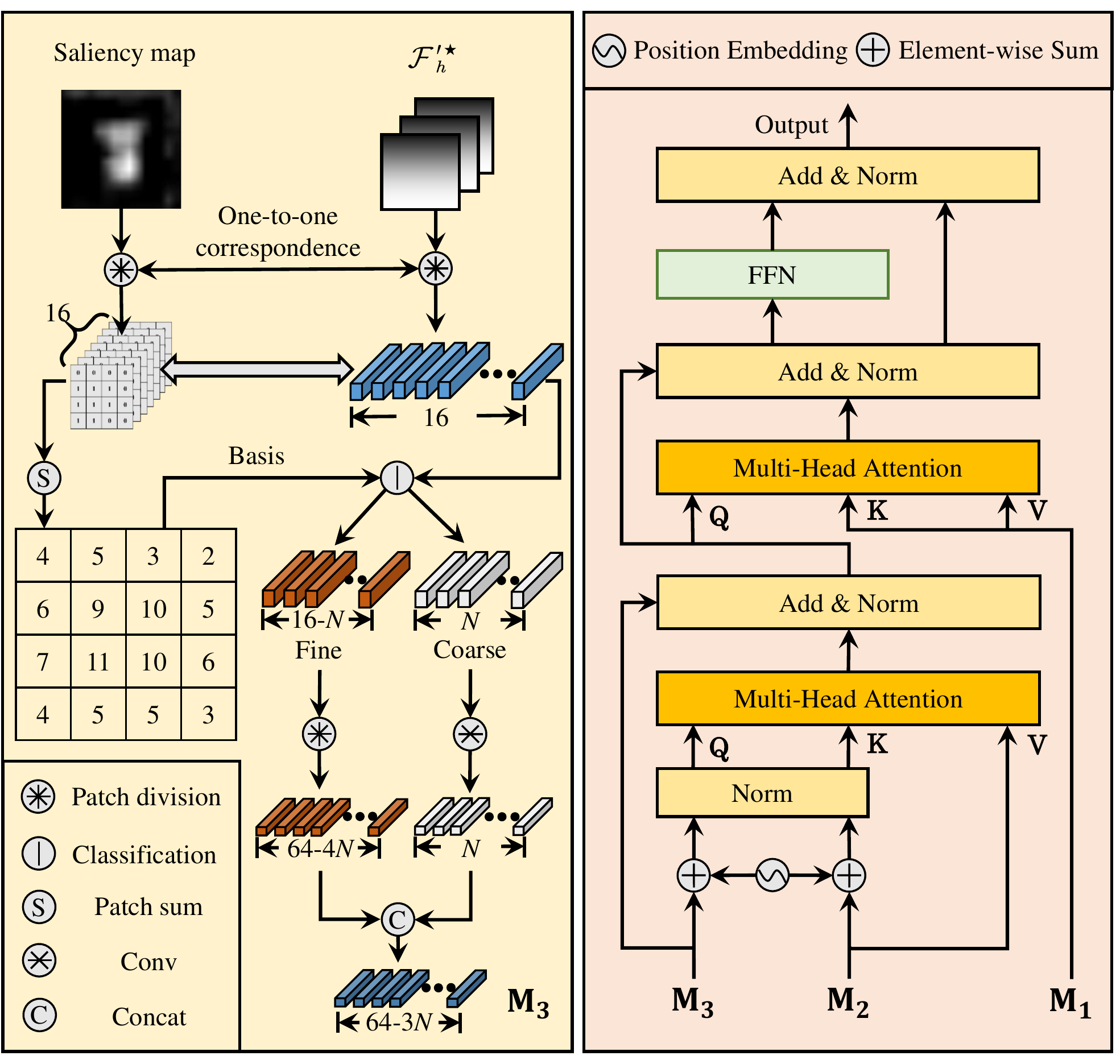}
	\caption{Detailed workflow of saliency adaption embedding operation (left) and saliency filtering Transformer (right). The $\mathcal{F}_{\rm{h}}^{  '\star}$ is the search region's features processed by feature adjustment sampling.}
	\label{fig:3}
\end{figure}

Based on the different attention on the foreground and background, not all patches possess equal information value for tracking tasks. As depicted in Fig.~\ref{fig:3}, a saliency adaption embedding operation is proposed to generate dynamic tokens for Transformer  based on initial saliency information. 

First, the thresholding method is used to process the $\mathbf{M}$ generated by the object saliency mining network. Considering the sampling to get binary mask is non-differentiable, the Gumbel-Softmax technique~\cite{jang2017categorical} is adopted:

\begin{equation}\label{2}
	\mathbf{P} = \mathrm{Gumbel-Softmax}(\mathbf{M})\in \{0,1\} \quad,
\end{equation}

The output $\mathbf{P}$ is a differentiable one-hot tensor, making it feasible for end-to-end training. Then, the entire saliency map and features are divided into patches using same-sized windows~\cite{liu2021swin}, ensuring that the mask and feature are aligned in spatial positions. Subsequently, each window is summed and the patches are separated into two parts according to the results. For parts with a higher value, the corresponding area may contain the object. Therefore the patches are further divided into smaller patches before being embedded into tokens. Conversely, the patches corresponding to small values are considered to contain a larger proportion of background information. Additionally, these patches are directly embedded into tokens. The tokens resulting from these patches, in combination with the tokens derived from the patches with higher values, are concatenated to form the input for the saliency filtering Transformer.

\Remark Different from the common dynamic token generation method to prune the useless tokens, the proposed approach feeds the entire tokens into the saliency fusion network at multiple levels. By distinguishing the contribution to the tracking task based on the saliency information, the Transformer can dynamically determine the significance of each token.  

\subsection{Saliency Filtering Transformer Network}
To refine the saliency information and enrich the appearance information, a new Transformer structure is designed. As illustrated in Fig.~\ref{fig:3}, the input to this network consists of three parts, template features $\mathbf{M}_1$ processed by feature adjustment sampling, dynamic tokens $\mathbf{M}_3$ generated by saliency adaption embedding and saliency features $\mathbf{M}_2$ obtained by object saliency mining.

For the encoder, different from the standard Transformer, the network uses the dynamic token $\mathbf{M}_{\rm3}$ as $\mathbf{Q}$ (Query), while the saliency features $\mathbf{M}_{\rm2}$ as $\mathbf{K}$ (Key) and $\mathbf{V}$ (Value). Generally, the scaled dot-product attention ($\mathrm{Att}$) can be calculated by as follows:
\begin{equation}\label{3}
	\mathrm{Att}(\mathbf{Q},\mathbf{K},\mathbf{V})= \mathrm{Softmax}(\frac{\mathbf{Q}\mathbf{K}^\top}{\sqrt{c}})\mathbf{V} \quad,
\end{equation}
where $\sqrt{c}$ is the scaling factor to avoid gradient vanishment in the softmax function. Then the procedure of the multi-head attention module $\mathrm{mAtt}$ is expressed as:
\begin{equation}\label{4}
		\begin{split}
			&\mathrm{mAtt}(\mathbf{Q},\mathbf{K},\mathbf{V}) = (\mathrm{Cat}(a^1,a^2,...,a^n))\mathbf{W}_{\rm c} \quad, \\
			&a^j = \mathrm{Att}(\mathbf{Q}\mathbf{W}_1^j,\mathbf{K}\mathbf{W}_2^j,\mathbf{V}\mathbf{W}_3^j) \quad,
		\end{split}
\end{equation}
where $\mathbf{W}_{\rm c} \in\mathbb{R}^{C \times C} $, $\mathbf{W}_{\rm 1}^j \in\mathbb{R}^{C \times C_{N}}$,$\mathbf{W}_{\rm 2}^j \in\mathbb{R}^{C \times C_{N}}$,$\mathbf{W}_{\rm 3}^j \in\mathbb{R}^{C \times C_{N}}$ can all be regarded as fully connected layer operation, where $C_{N}=C/N$, and $N$ is the number of parallel attention head. Afterward, the output of the encoder $\mathbf{M}_{\rm4}$ can be obtained by the scaled dot-product attention ($\mathrm{Att}$) can be calculated by as follows:
\begin{equation}\label{5}
	\begin{split}
	&\mathbf{M}_{\rm x} = \mathrm{mAtt}(\mathbf{M}_{\rm3},\mathbf{M}_{\rm2},\mathbf{M}_{\rm2}) \quad,\\
	&\mathbf{M}_{\rm4}=\mathrm{Norm}(\mathbf{M}_{\rm x}+\mathbf{M}_{\rm3}) \quad, 
	\end{split}
\end{equation}

\noindent where $\mathbf{M}_{\rm x}$ is the intermediate variable and $\mathrm{Norm}$ is the normalization operation. Considering the network's lightweight, we removed FFN module from the encoder. 

\Remark By leveraging the salience features as prior information, the model gives greater emphasis to the salient regions as identified through object salience mining. 

To further refine the saliency information while mitigating the loss of appearance information, the decoder is structured as a standard Transformer that takes the features of the objeact in the template as input. The network uses the features $\mathbf{M}_{\rm1}$ as $\mathbf{K}$ and $\mathbf{V}$, and the output of the encoder $\mathbf{M}_{\rm4}$ as $\mathbf{Q}$. The overall computing mechanism is the same as the encoder. The output of the decoder $\mathbf{M}_{\rm5}$ can be derived by:
\begin{equation}\label{6}
	\begin{split}
		&\mathbf{M}_{\rm y} = \mathrm{mAtt}(\mathbf{M}_{\rm4},\mathbf{M}_{\rm1},\mathbf{M}_{\rm1})\quad\quad, \\
		&\mathbf{M}_{\rm5}= \mathrm{Norm}(\mathbf{M}_{\rm y}+\mathbf{M}_{\rm4})\quad,
	\end{split}	
\end{equation} 

\noindent where $\mathbf{M}_{\rm y}$ is the intermediate variable.

\begin{figure*}[!t]	
	\raggedright
	\includegraphics[width=0.325\linewidth]{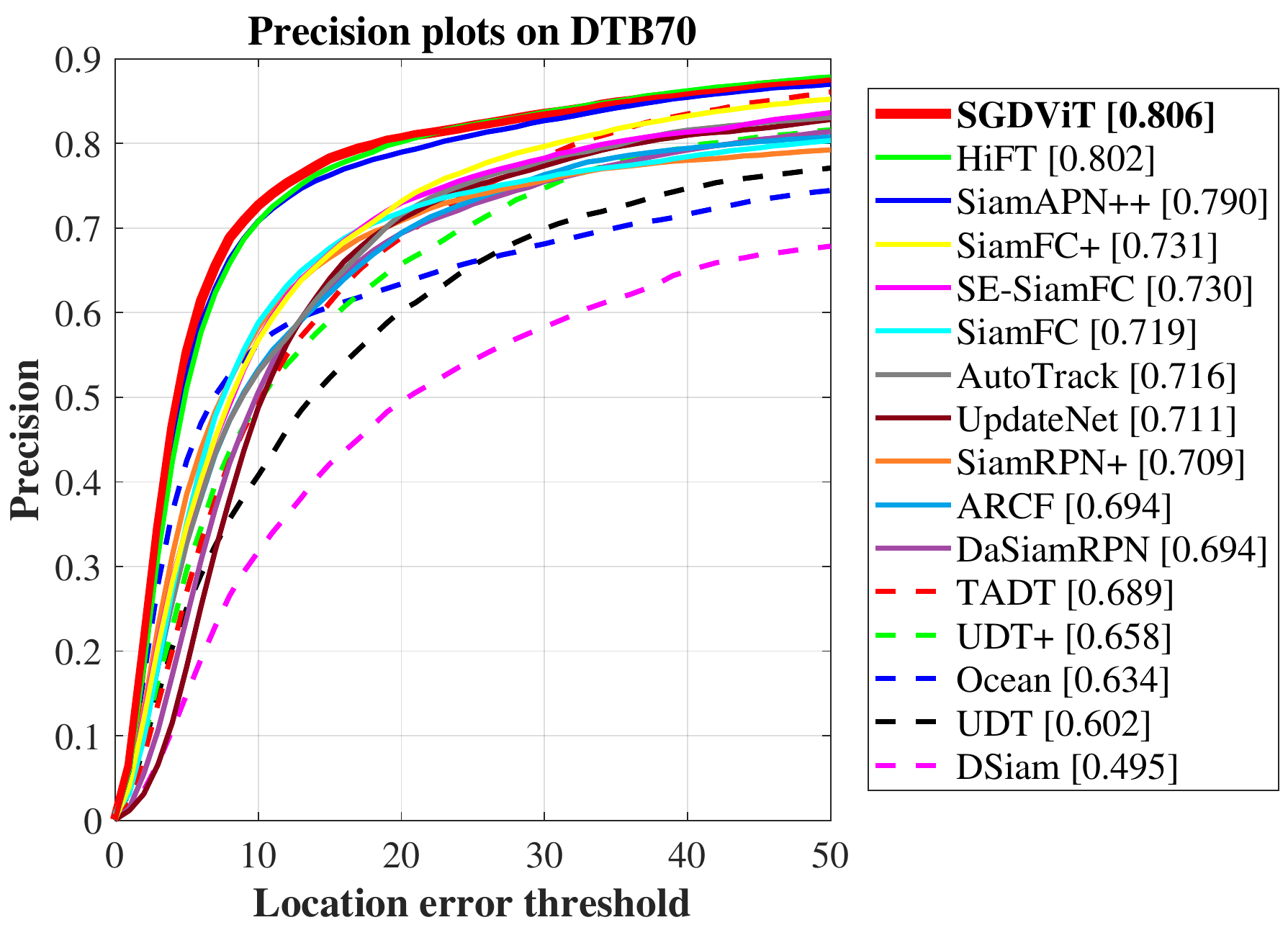}
	\includegraphics[width=0.325\linewidth]{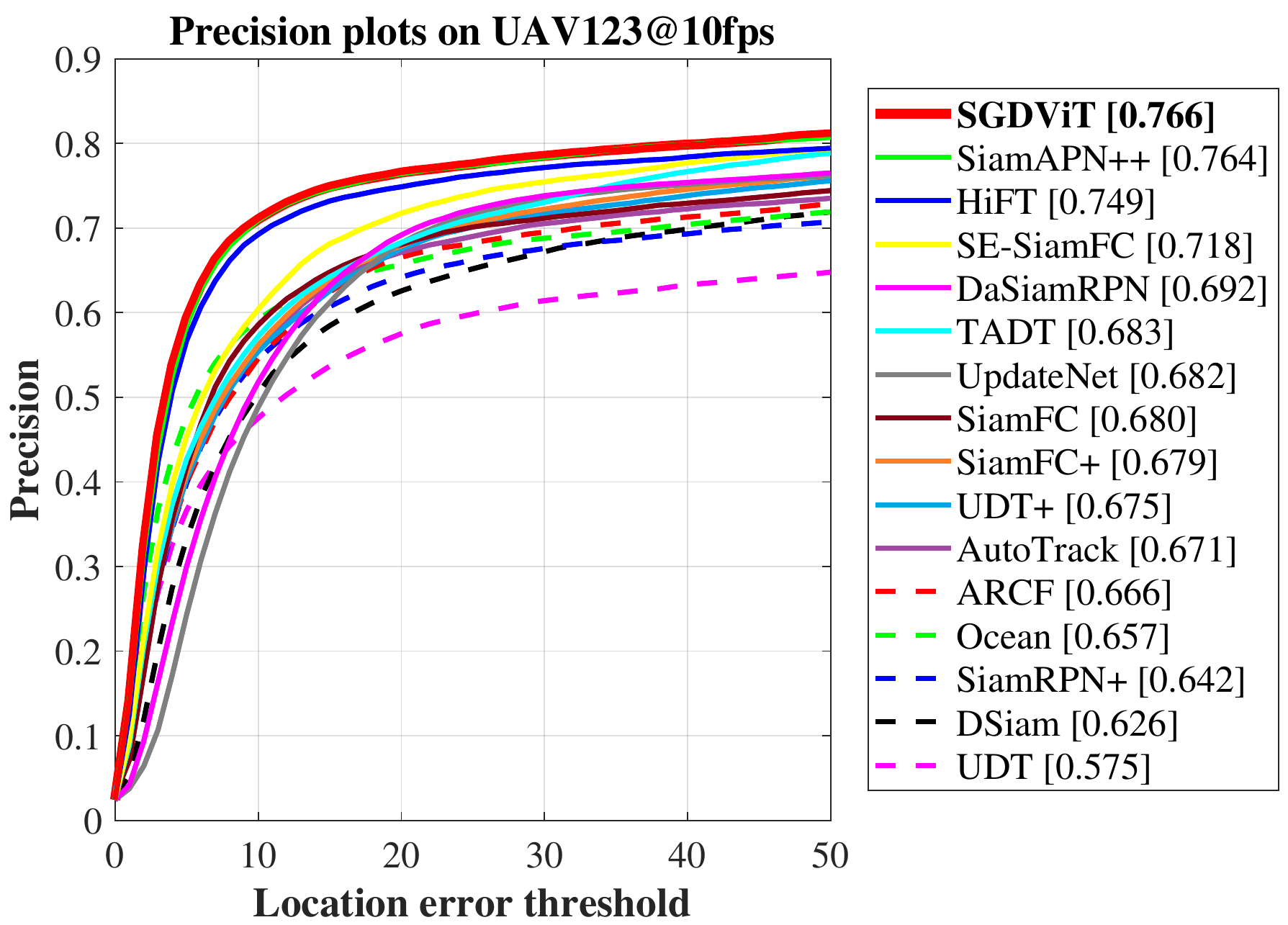}
	\includegraphics[width=0.325\linewidth]{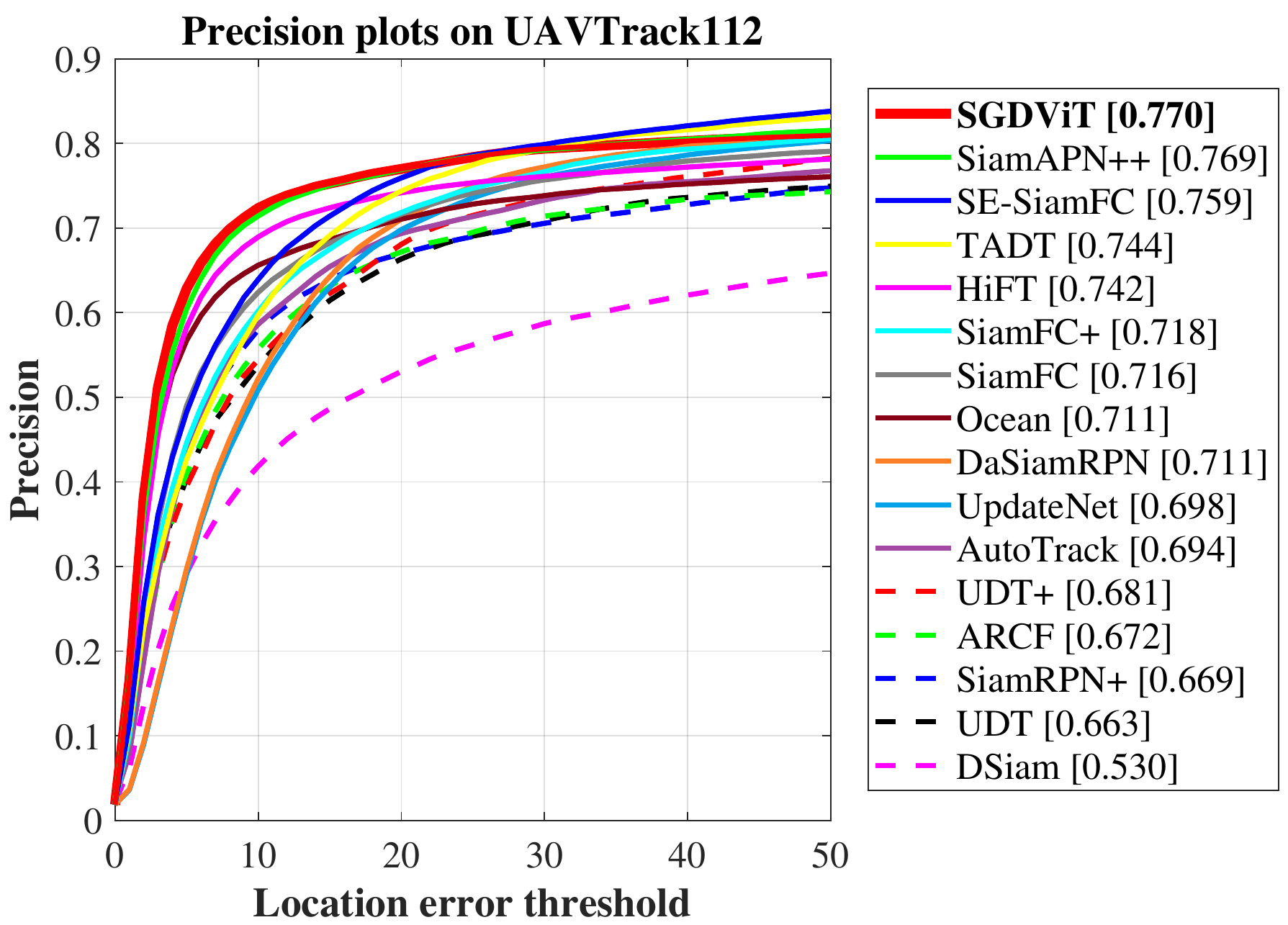}
    \label{fig:all-1}
    \vspace{-0.3cm}
\end{figure*}
\begin{figure*}[!t]	
	\raggedright
	\includegraphics[width=0.325\linewidth]{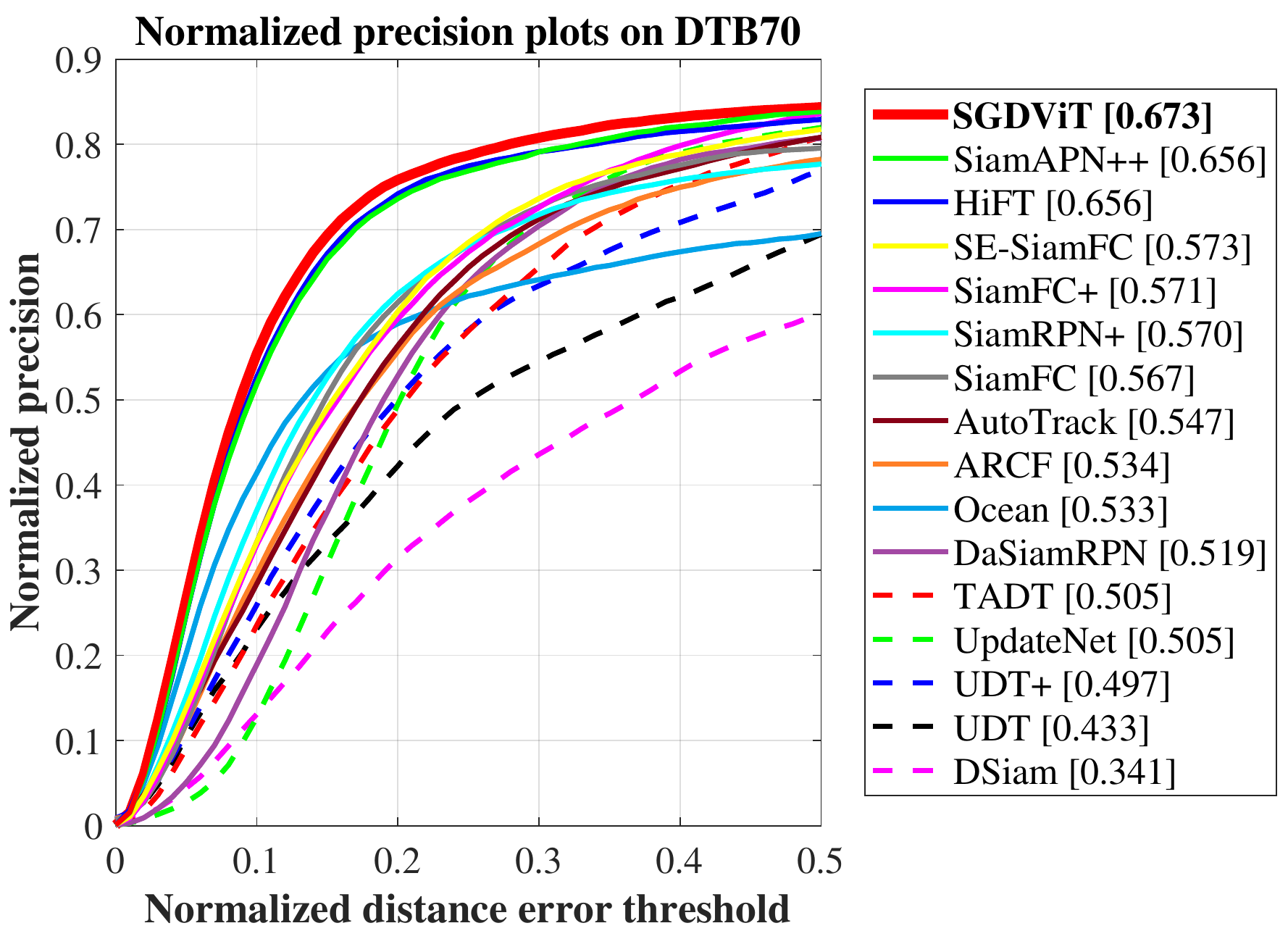}
	\includegraphics[width=0.325\linewidth]{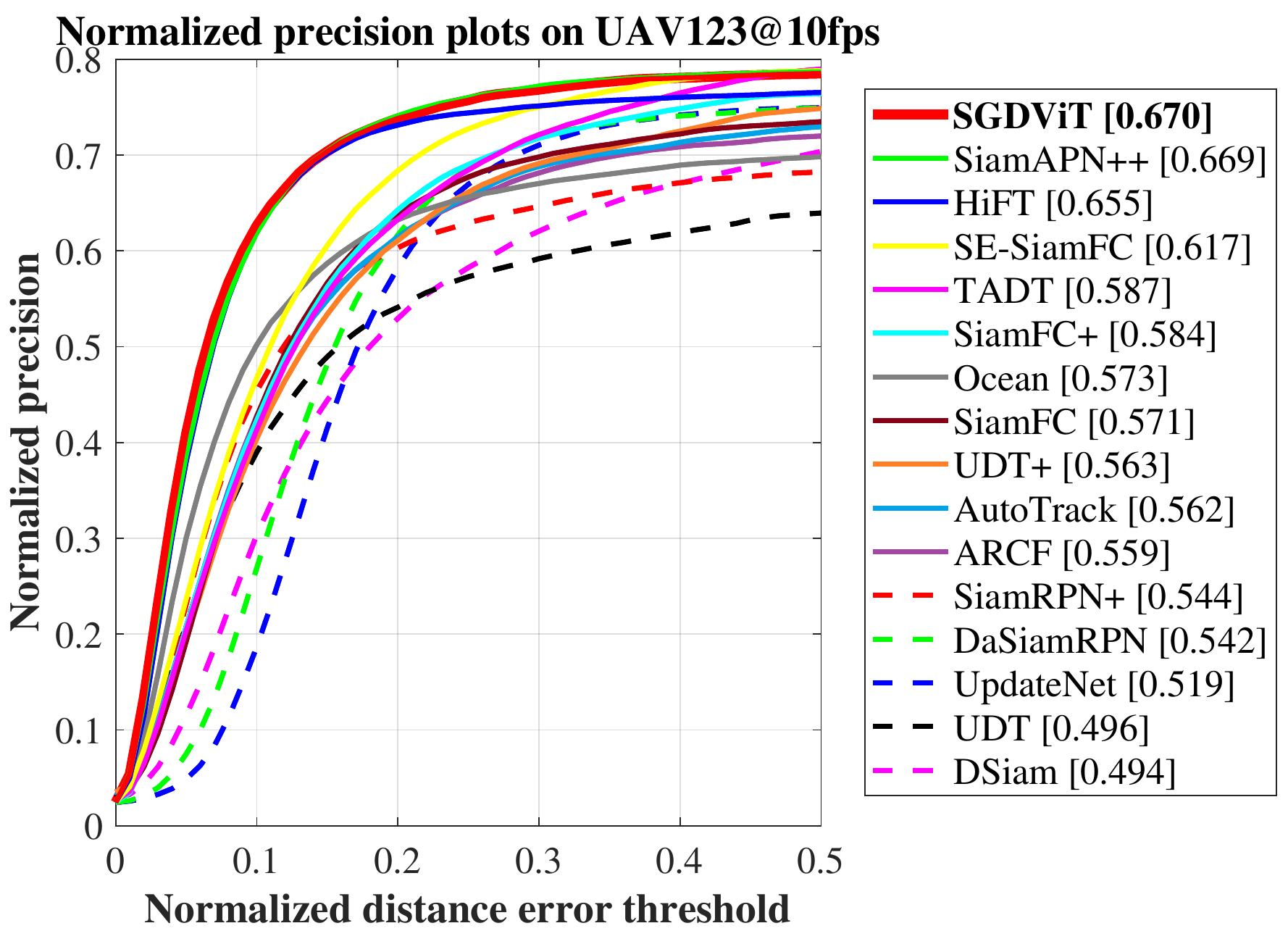}
	\includegraphics[width=0.325\linewidth]{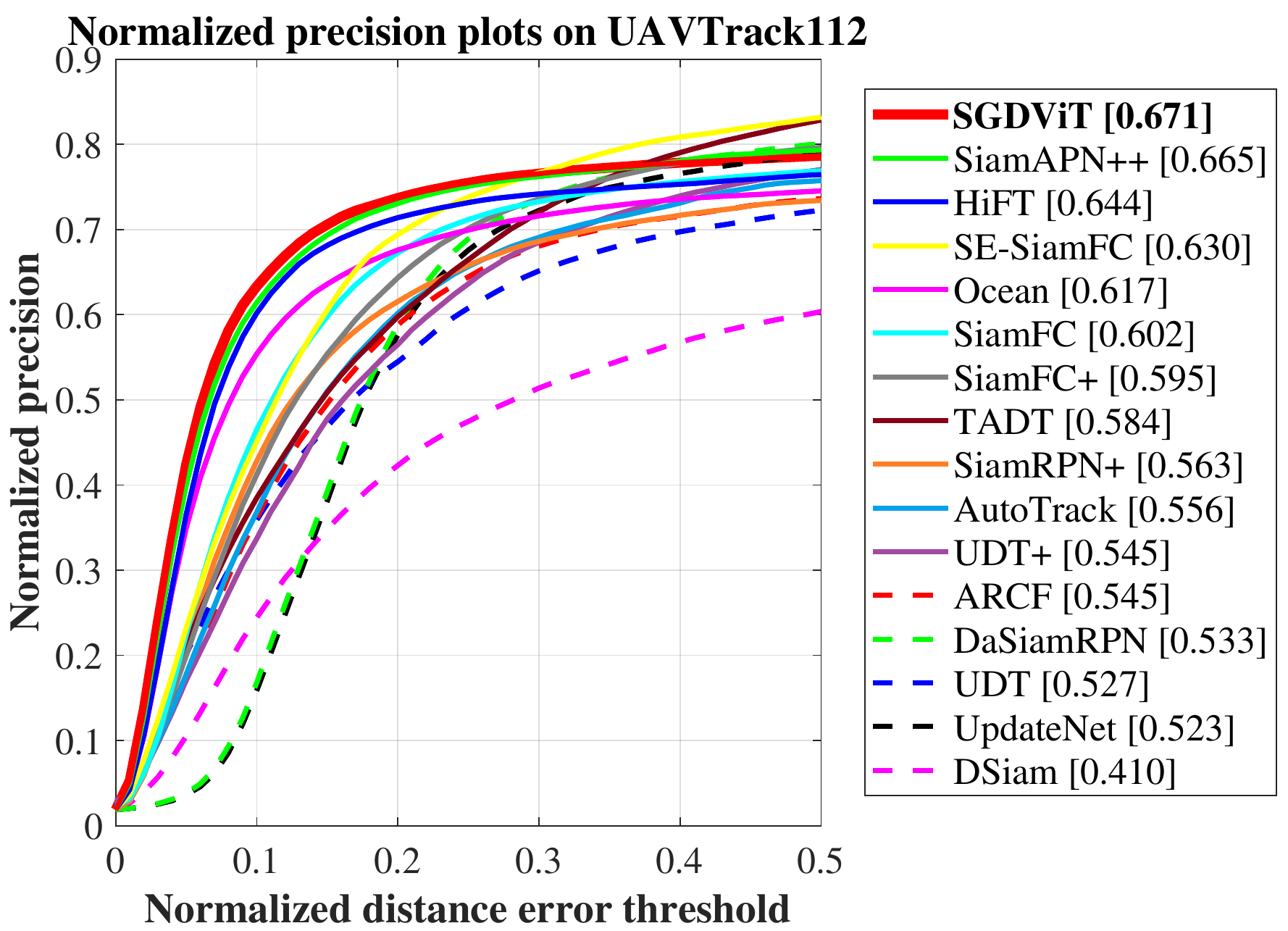}
    \label{fig:all-1}
    \vspace{-0.4cm}
\end{figure*}
\begin{figure*}[!t]	
	\includegraphics[width=0.325\linewidth]{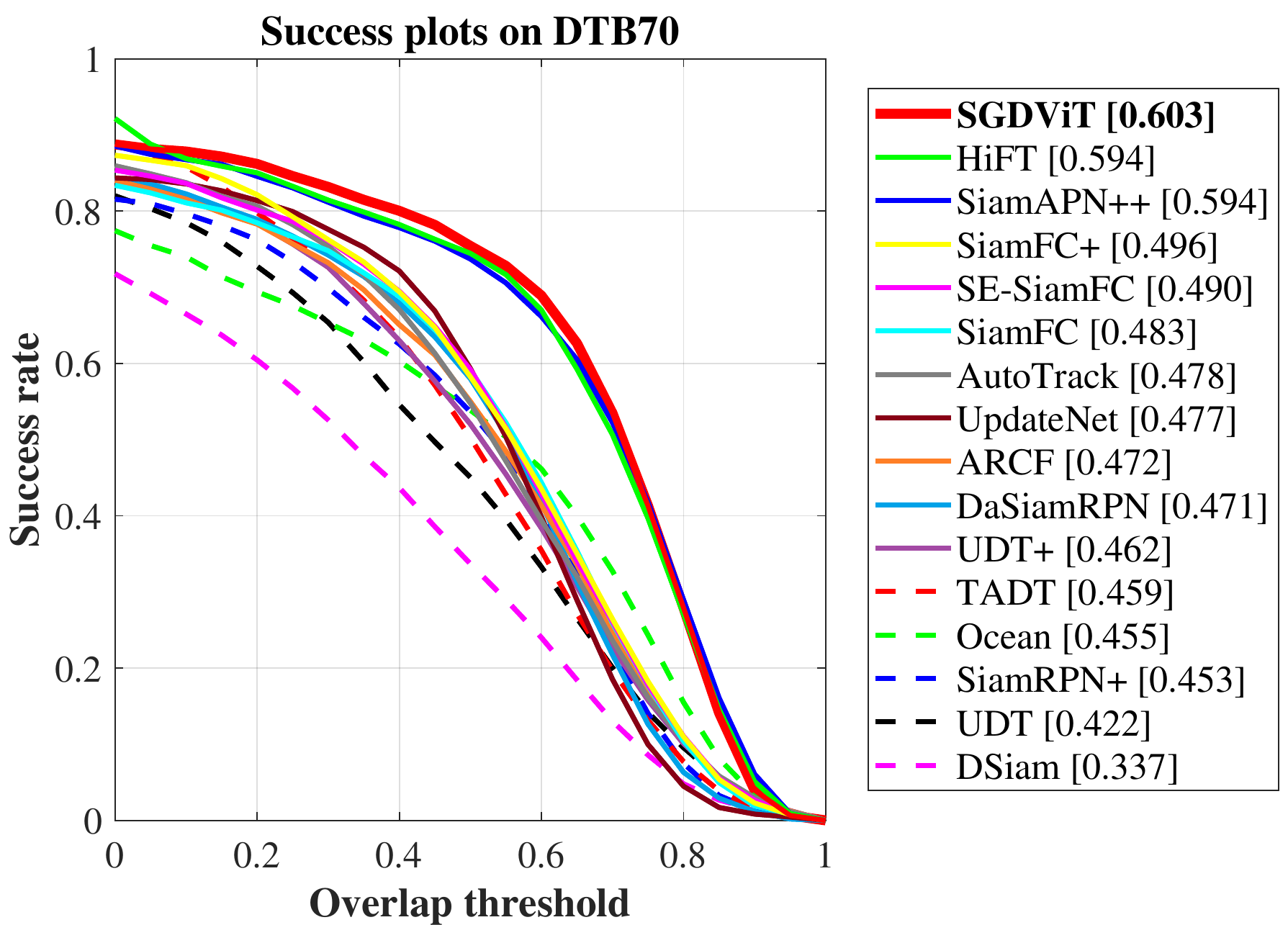}
	\includegraphics[width=0.325\linewidth]{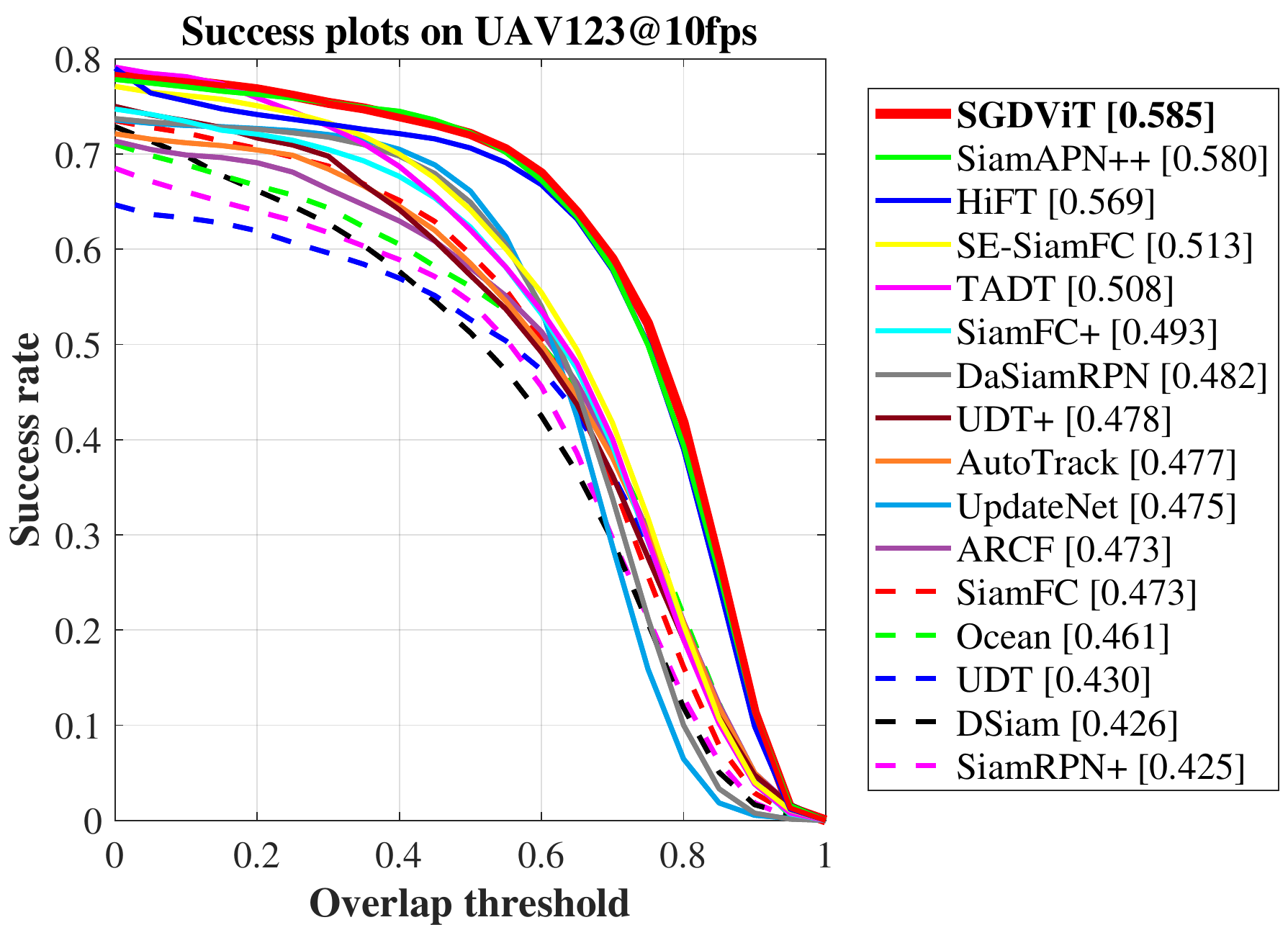}
	\includegraphics[width=0.325\linewidth]{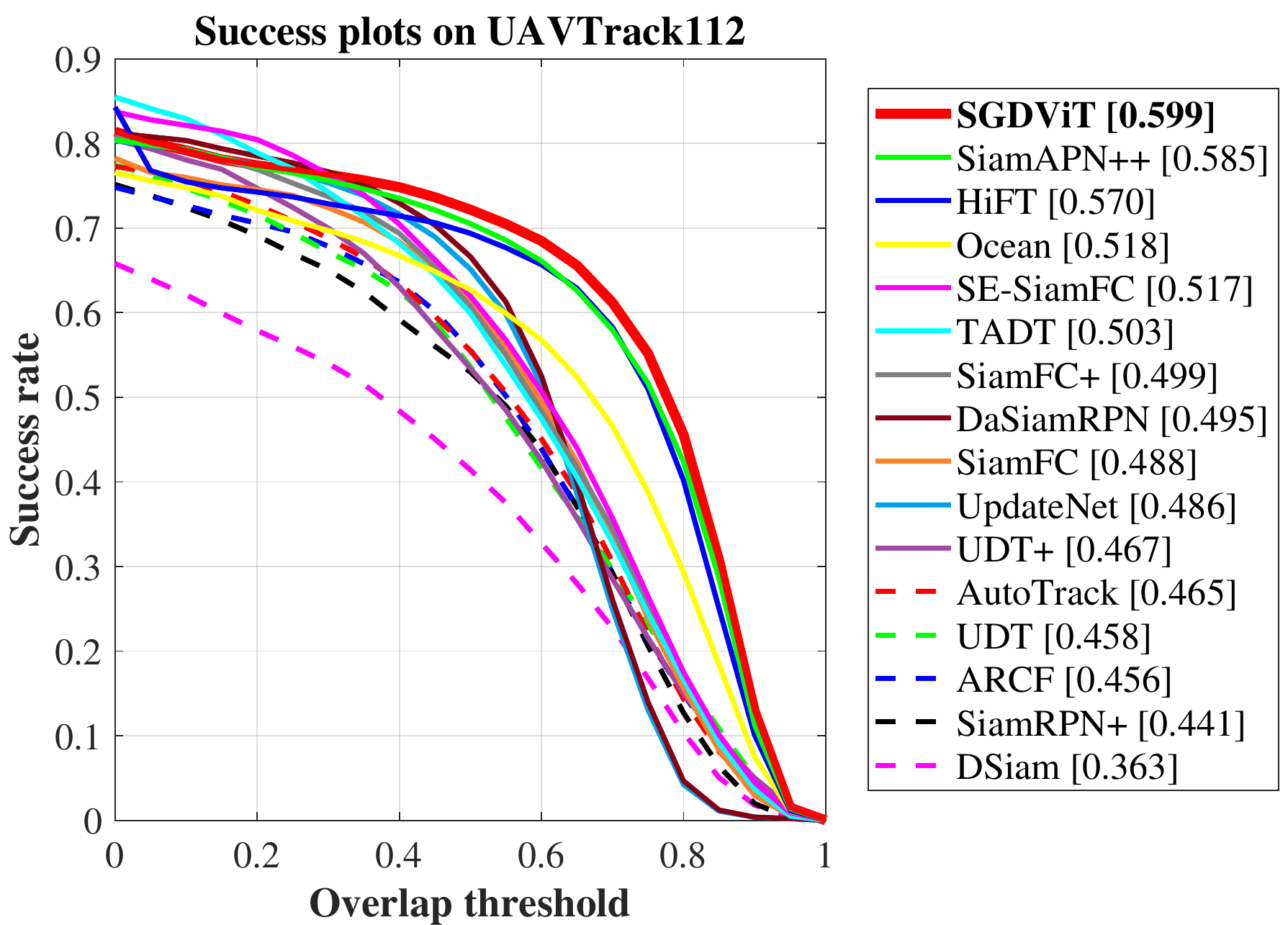}
	\setlength{\abovecaptionskip}{-2pt} 
	\caption
	{
		 Overall performance of SGDViT and other SOTA trackers on DTB70 \cite{li2017visual} (the first column), UAV123@10fp \cite{mueller2016benchmark} (the second column), and UAVTrack112 \cite{fu2021onboard} (the third column) benchmarks. The experimental results demonstrate that the proposed method yields superior performance on all benchmarks.
	}
	\label{fig:all}
	\vspace{-0.5cm}
\end{figure*}

After a feed-forward network (FFN) and normalization (Norm), the output $\mathbf{M}_{\rm o}$ can be formulated as:  
\begin{equation}\label{7}
	\mathbf{M}_{\rm o}= \mathrm{Norm}(\mathrm{FFN}(\mathbf{M}_{\rm5})+\mathbf{M}_{\rm5})\quad. 
\end{equation} 

\Remark Using the template features as a guidance for the decoding process, the network is able to concentrate on the salient features of the object and minimize the residual impurities present in the prior knowledge. The combination of encoder and decoder enables similarity information to undergo a progressive refinement process from coarse to fine, thereby enhancing the efficacy of the SGDViT tracker.


\section{Experiment}

\subsection{Implementation Details}
The proposed tracker SGDViT is trained for 100 epochs on a PC equipped with an Intel \emph{i}9-9920X CPU, 32GB of RAM, and 2 NVIDIA TITAN RTX GPUs. During the training progress, the last three layers of AlexNet \cite{krizhevsky2017imagenet} are fine-tuned in the last 90 epochs while the first two layers are frozen. The overall loss function is determined the same as HiFT\cite{cao2021hift}. The sizes of template $\mathbf{Z}$ and search region $\mathbf{X}$ are respectively set to $127\times127$ and $287\times287$. Furthermore, the learning rate is initialized as $5\times10^{-4}$ and then decreased in the log space from $10^{-2}$ to $10^{-4}$. Additionally, the image pairs are extracted from train benchmarks including COCO \cite{lin2014microsoft}, ImageNet VID \cite{russakovsky2015imagenet}, GOT-10K \cite{huang2019got}, and Youtube-BB \cite{real2017youtube}. 

\subsection{Overall Performance on UAV Tracking Benchmarks}
The three metrics in the one-pass evaluation (OPE) metrics~\cite{muller2018trackingnet} are precision, normalized precision, and success rate. The center location error (CLE) determines the precision. The percentage of frames with a lower CLE than 20 pixels is presented as the precision plot (PP). The intersection over
union (IoU) is adopted to measure the success rate. The
percentage of frames that have a larger IoU than the preset
maximum threshold is reported as the success plot (SP). The area-under-the-curve (AUC) on SP is used
to rank the success rate of trackers. Considering that the precision metric can be affected by the image resolution and the bounding box scale, the normalized precision metric is adopted. In this section, SGDViT is tested with other 15 state-of-the-art (SOTA) trackers on three authoritative UAV tracking benchmarks, including HiFT~\cite{cao2021hift}, SiamAPN++~\cite{cao2021siamapn++}, SiamFC~\cite{bertinetto2016fully}, SiamRPN+~\cite{zhang2019deeper}, SiamFC+~\cite{zhang2019deeper}, Ocean~\cite{zhang2020ocean}, DaSiamRPN~\cite{zhu2018distractor}, SE-SiamFC~\cite{sosnovik2021scale}, AutoTrack~\cite{li2020autotrack}, ARCF~\cite{huang2019learning}, TADT~\cite{li2019target}, UDT+~\cite{wang2019unsupervised}, UDT~\cite{wang2019unsupervised}, Dsiam~\cite{guo2017learning}, and UpdateNet~\cite{zhang2019learning}.  

\textbf{DTB70}~\cite{li2017visual}: DTB70 has 70 difficult UAV sequences, most of which are vigorous motion situations. As shown in the first column of Fig.~\ref{fig:all}, 
SGDViT can obtain satisfactory performance, achieving the best precision (0.806), normalized precision (0.673), and overall success score (0.603).

\textbf{UAV123@10fps} \cite{mueller2016benchmark}: UAV123@10fps is obtained by downsampling from the original 30fps version. Hence, the motion problems in UAV123@10fps are worse than in UAV123. SGDViT can consistently achieve excellent results, as shown in the second column of Fig.~\ref{fig:all}, attaining the highest success rate (0.585), precision (0.766), and normalized precision (0.670). 

\textbf{UAVTrack112} \cite{fu2021onboard}: UAVTrack112 records a wide range of challenging aerial tracking scenarios. As shown in Fig.~\ref{fig:all}, SGDViT ranks first in precision (0.770), normalized precision (0.671), and success rate (0.599).


	


\subsection{Attribute-Based Comparison}
To provide a more comprehensive analysis of the robustness of SGDViT, the tracker's performance was evaluated on various aerial-specific attributes, including aspect ratio change, scale variation, low resolution, and background clutter, as shown in Tab.~\ref{tab:1}. Our tracker is endowed with the ability to track objects with aspect ratio change, scale variation, and low resolution since the saliency filtering Transformer fuses the information further with the guidance of saliency information.

\begin{table*}[!t]
 \centering
 \caption{Qualitative attribute-based comparisons of top 6 trackers on three benchmarks. The best two performances are respectively highlighted by \textbf{bold} and \underline{underlined}. SGDViT keeps achieving the best performance in different attributes.}
    \label{table1}
\setlength{\tabcolsep}{5mm}
  \begin{tabular}{lccccccccc|c}
   \hline
   \multirow{2}{*}{Attributes} & \multicolumn{2}{c}{Aspect ratio change} & \multicolumn{2}{c}{Scale variation} & \multicolumn{2}{c}{Low resolution} & \multicolumn{2}{c}{Background clutter} \\
  \cline{2-9} &Prec. &  Succ. & Prec. &  Succ.& Prec. &  Succ. & Prec. &  Succ.\\
  \hline 
   HiFT  & 0.696 & 0.518 & 0.725 & 0.549 & 0.551 & 0.365 & \underline{0.784} & \underline{0.567} \\ 
   SiamAPN++  & \underline{0.717} & \underline{0.537} & \underline{0.736}  & \underline{0.556} & \underline{0.595} & \underline{0.395} & 0.728 & 0.520 \\ 
   SE-SiamFC  & 0.668 & 0.461 & 0.682 & 0.483 & 0.552 & 0.323 & 0.662 & 0.398 \\ 
   SiamFC  & 0.626 & 0.414 & 0.649 & 0.444 & 0.549 & 0.317 & 0.624 & 0.391 \\ 
   Ocean  & 0.610 & 0.414 & 0.630 & 0.436 & 0.583 & 0.381 & 0.553 & 0.362 \\ 
   \midrule
   \textbf{SGDViT (Ours)}  & \textbf{0.731} & \textbf{0.549} & \textbf{0.745} & \textbf{0.567} & \textbf{0.617} & \textbf{0.410} & \textbf{0.805} & \textbf{0.580} \\ 
   \hline 
      \end{tabular}
      \label{tab:1}%
\end{table*}


\subsection{Ablation Study}
\begin{figure}[!t]	
	\centering
	\includegraphics[width=1\linewidth]{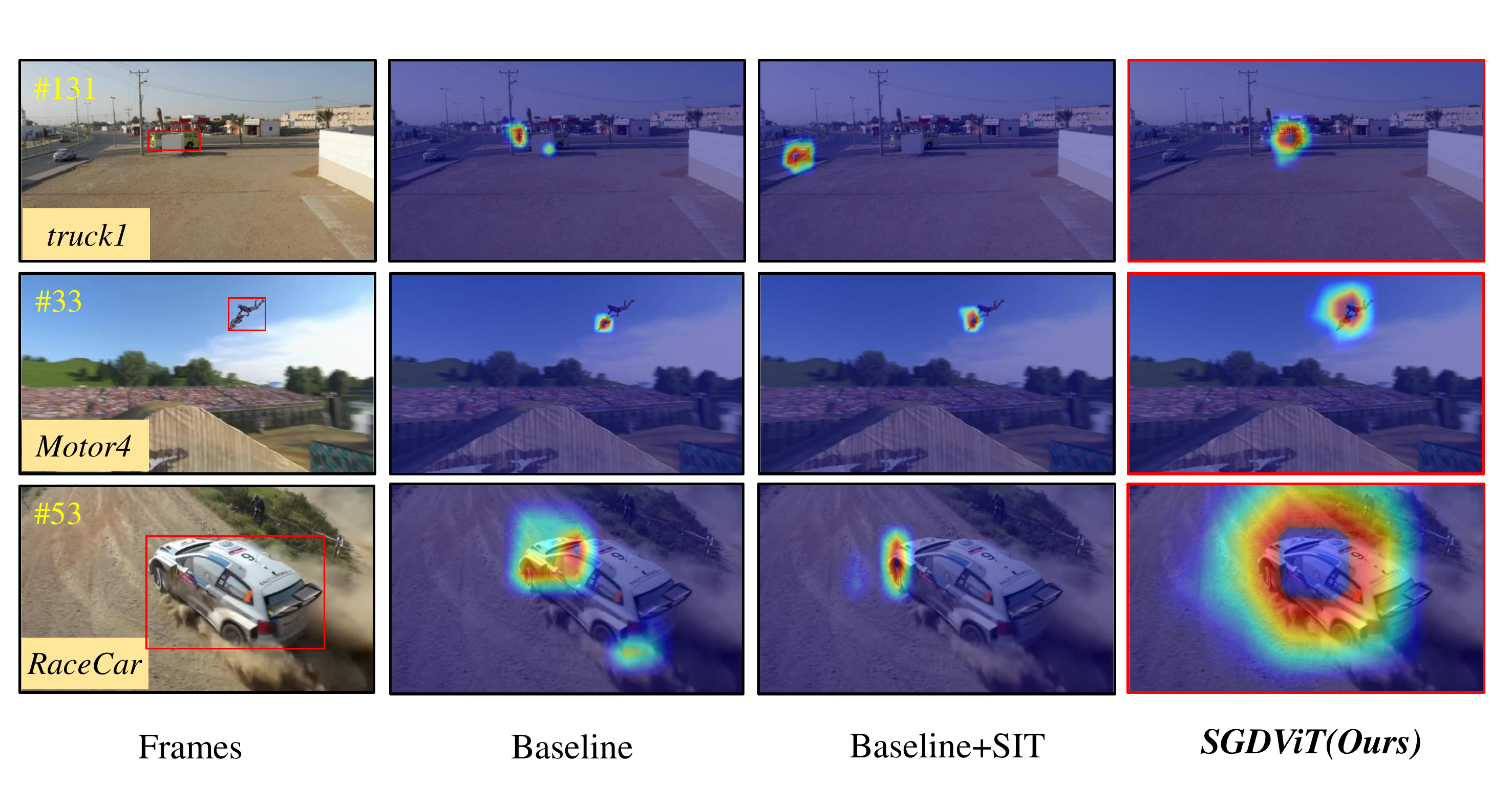}
	\setlength{\abovecaptionskip}{-0.5cm}
	\caption
	{
        Visualization of the confidence map of three tracking methods on several sequences from UAV123@10fps \cite{mueller2016benchmark} and
        DTB70 \cite{li2017visual}. The target objects are marked out by \textcolor[rgb]{1,0,0}{red} boxes in the original frames. 
	}
	\label{fig:heat}
\end{figure}

In this section, the contribution of each module is analyzed through experiments conducted on DTB70.
This work considers baseline as the model with only a feature extraction network and classification \& regression network. SIT denotes utilizing the similarity map as the input of the Transformer directly. SAT indicates using the saliency features but without the dynamic token generation. SGDViT denotes the full version of the proposed saliency-guided dynamic vision Transformer. In Tab.~\ref{tab:2}, the performance improvement of adding the original Transformer with similarity map (Baseline+SIT) is not high, only 1.4\% on precision and 1.8\% on success rate. Replacing the similarity map with saliency features, Baseline+SAT raises the success rate by 7.8\%. Baseline+SAT+Dyn (SGDViT) obtains the best performance, enhancing both the precision and success rate by 9.6\%. As shown in Fig.~\ref{fig:heat}, the confidence map of our SGDViT tracker consistently focuses on the object under various challenges in aerial tracking, \textit{e.g.}, scale variation and occlusion in \emph{truck1}, aspect ratio change and scale variation in \emph{Motor4}, and appearance change in \emph{RaceCar}.

\begin{table}[!t]
    \centering
    
    \caption{Qualitative comparisons between different combinations of modules in SGDViT framework on DTB70. $\Delta$ denotes the variation of the metrics. The best performances are highlighted by \textbf{bold}. 
	}
	\setlength{\tabcolsep}{1mm}
    \begin{tabular}{ccccc}
         \toprule
         \multicolumn{1}{c}{Trackers} & \multicolumn{1}{c}{Prec.} &\multicolumn{1}{c}{$\Delta_{prec}$(\%)}
         &\multicolumn{1}{c}{Succ.}&\multicolumn{1}{c}{$\Delta_{succ}$(\%)}\\
         \midrule
         \multicolumn{1}{c}{Baseline}& 0.735 & - & 0.550 & -\\
         \multicolumn{1}{c}{Baseline+SIT}& 0.746 & +1.4 & 0.560 & +1.8\\
         \multicolumn{1}{c}{Baseline+SAT}& 0.779 & +5.9 & 0.593 & +7.8\\
         \midrule
         \multicolumn{1}{c}{\textbf{Baseline+SAT+Dyn (SGDViT)}}&\textbf{0.806} & \textbf{+9.6} &\textbf{0.603} & \textbf{+9.6}\\
         
         \bottomrule
    \end{tabular}
    \label{tab:2}%
\end{table}

\section{Real-World Tests}
\label{sec:Real-WorldTests}
\begin{figure}[!t]	
	\centering
	\includegraphics[width=1\linewidth]{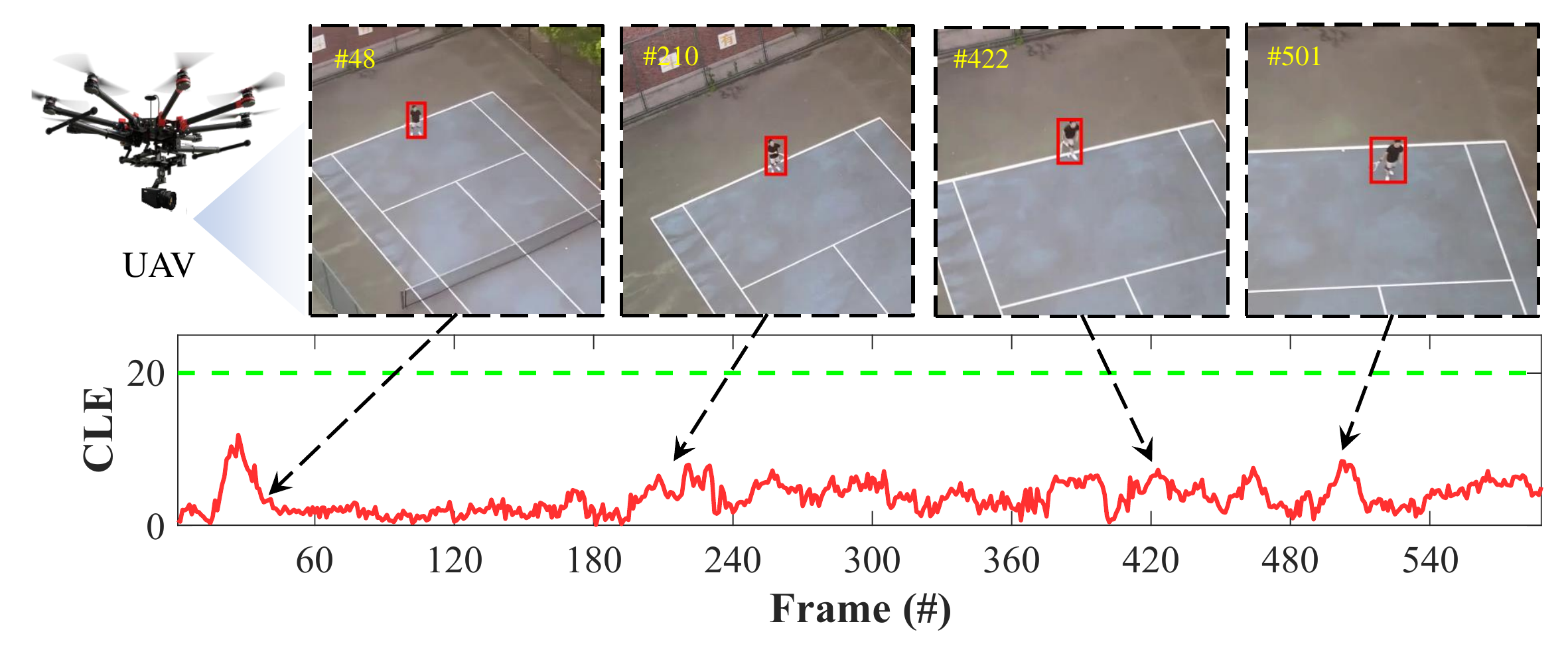}
	\includegraphics[width=1\linewidth]{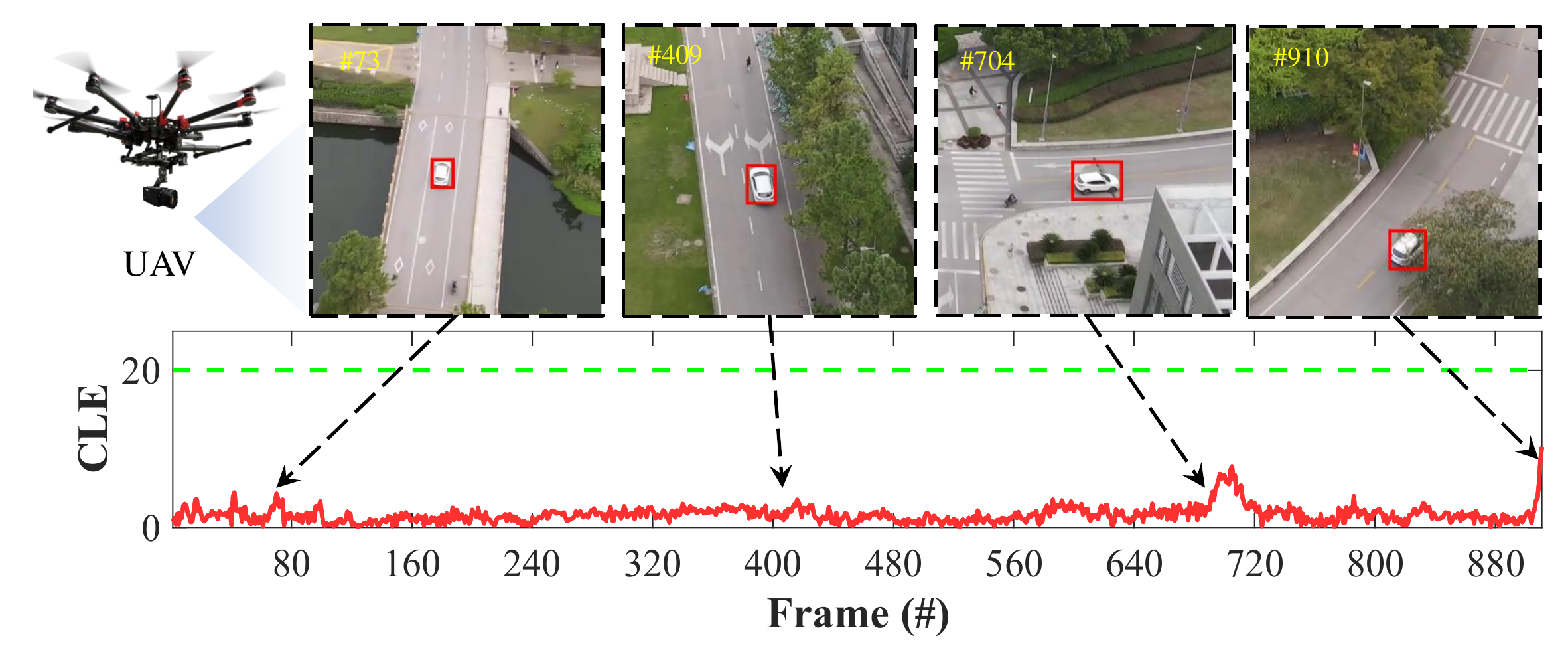}
	\includegraphics[width=1\linewidth]{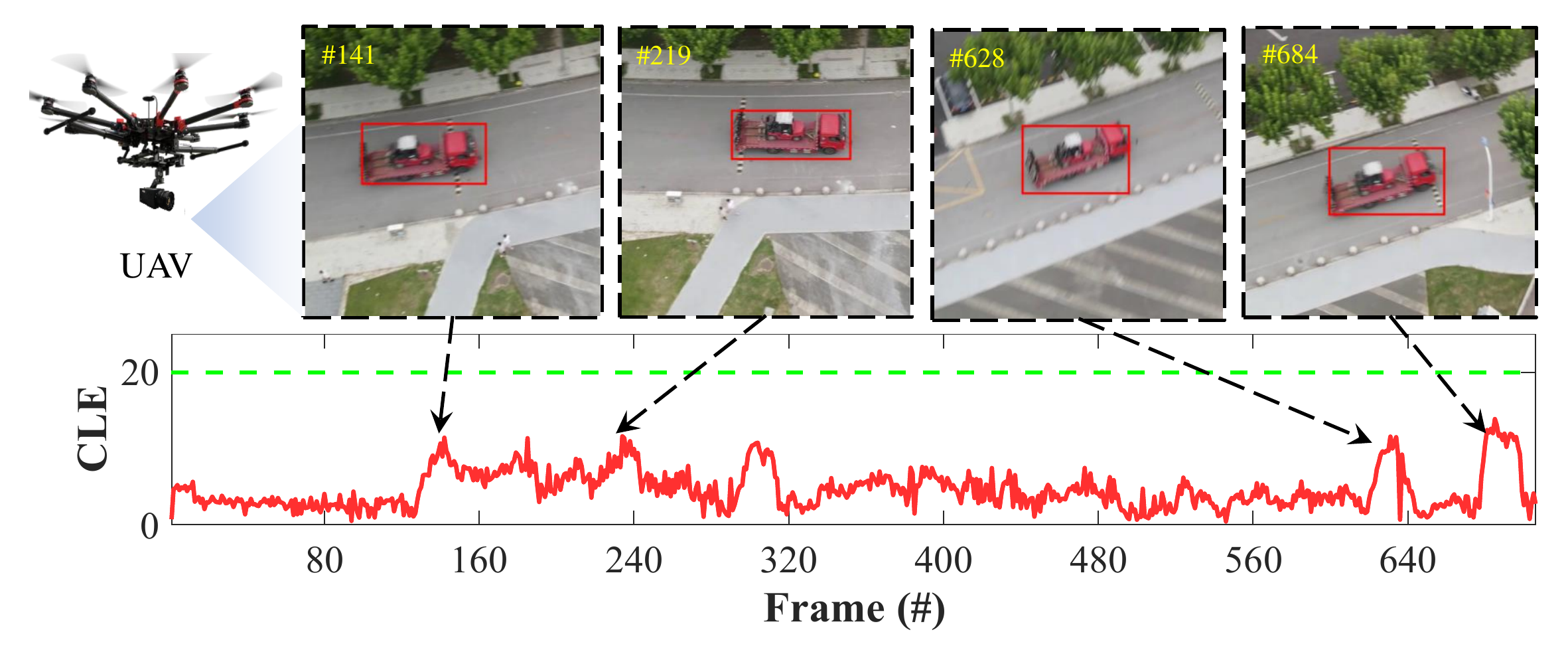}
	\setlength{\abovecaptionskip}{-0.4cm}
	\caption
	{
         Real-world tests on a typical UAV platform. SGDViT achieves robust performance with 23 frames per second. The \textcolor[rgb]{1,0,0}{red} bounding boxes represent the tracking results. The CLE below the \textcolor[rgb]{0,1,0}{green} dashed
line is the success tracking in the real-world test.}
	\label{fig:real}
\end{figure}

Extensive real-world tests are conducted and three of them are presented in Fig.~\ref{fig:real}. Specifically, SGDViT is tested on a typical UAV platform equipped with an NVIDIA Jetson AGX Xavier. In Test 1, the primary challenge faced is a significant change in viewpoint, resulting in a constantly evolving shape of the target object. In Test 2, the appearance of the object in the images undergoes substantial changes, and occlusions from time to time exacerbates the difficulties associated with robust tracking. Test 3 presents a dual challenge, with both the scale and shape of the target object undergoing variations. Thus, the real-world tests validate SGDViT’s robustness and efficiency in a variety of UAV-specific scenarios.


\section{Conclusions}
\label{sec:Conclusions}
In this work, a novel saliency-guided dynamic vision Transformer for UAV tracking is proposed. This method revolutionizes the information fusion method. The object saliency mining network extracts saliency information through the integration of spatial and channel information fusion. The saliency filtering Transformer refines the saliency information and enriches the appearance information. Additionally, a new saliency adaption embedding operation is proposed to accelerate the Transformer structure. Abundant experimental results prove the effectiveness of our tracker. To sum up, we firmly believe that the proposed approach will contribute to the advancement of UAV tracking.


\section*{Acknowledgment}
This work is supported by the Natural Science Foundation of Shanghai (No. 20ZR1460100) and the National Natural Science Foundation of China (No. 62173249). 

\bibliographystyle{IEEEtran}
\balance
\bibliography{yll_ICRA}

\begin{thebibliography}{10}
\providecommand{\url}[1]{#1}
\csname url@samestyle\endcsname
\providecommand{\newblock}{\relax}
\providecommand{\bibinfo}[2]{#2}
\providecommand{\BIBentrySTDinterwordspacing}{\spaceskip=0pt\relax}
\providecommand{\BIBentryALTinterwordstretchfactor}{4}
\providecommand{\BIBentryALTinterwordspacing}{\spaceskip=\fontdimen2\font plus
\BIBentryALTinterwordstretchfactor\fontdimen3\font minus
  \fontdimen4\font\relax}
\providecommand{\BIBforeignlanguage}[2]{{%
\expandafter\ifx\csname l@#1\endcsname\relax
\typeout{** WARNING: IEEEtran.bst: No hyphenation pattern has been}%
\typeout{** loaded for the language `#1'. Using the pattern for}%
\typeout{** the default language instead.}%
\else
\language=\csname l@#1\endcsname
\fi
#2}}
\providecommand{\BIBdecl}{\relax}
\BIBdecl

\bibitem{zhang2019eye}
H.~Zhang, G.~Wang, Z.~Lei, and J.-N. Hwang, ``{Eye in the Sky: Drone-Based
  Object Tracking and 3D Localization},'' in \emph{Proceedings of the ACM
  International Conference on Multimedia}, 2019, pp. 899--907.

\bibitem{bonatti2019IROS}
R.~Bonatti, C.~Ho, W.~Wang, S.~Choudhury, and S.~Scherer, ``{ Towards a Robust
  Aerial Cinematography Platform: Localizing and Tracking Moving Targets in
  Unstructured Environments},'' in \emph{Proceedings of the IEEE/RSJ
  International Conference on Intelligent Robots and Systems (IROS)}, 2019, pp.
  229--236.

\bibitem{Ye_2021_TIE}
J.~Ye, C.~Fu, F.~Lin, F.~Ding, S.~An, and G.~Lu, ``{Multi-Regularized
  Correlation Filter for UAV Tracking and Self-Localization},'' \emph{IEEE
  Transactions on Industrial Electronics}, vol.~69, no.~6, pp. 6004--6014,
  2022.

\bibitem{fu2022siamese}
C.~Fu, K.~Lu, G.~Zheng, J.~Ye, Z.~Cao, B.~Li, and L.~Geng, ``{Siamese Object
  Tracking for Unmanned Aerial Vehicle: A Review and Comprehensive Analysis},''
  \emph{arXiv preprint arXiv:2205.04281}, pp. 1--33, 2022.

\bibitem{li2019siamrpn++}
B.~Li, W.~Wu, Q.~Wang, F.~Zhang, J.~Xing, and J.~Yan, ``{SiamRPN++: Evolution
  of Siamese Visual Tracking with very Deep Networks},'' in \emph{Proceedings
  of the IEEE/CVF Conference on Computer Vision and Pattern Recognition
  (CVPR)}, 2019, pp. 4282--4291.

\bibitem{xu2020siamfc++}
Y.~Xu, Z.~Wang, Z.~Li, Y.~Yuan, and G.~Yu, ``{SiamFC++: Towards Robust and
  Accurate Visual Tracking with Target Estimation Guidelines},'' in
  \emph{Proceedings of the AAAI Conference on Artificial Intelligence (AAAI)},
  vol.~34, no.~07, 2020, pp. 12\,549--12\,556.

\bibitem{cao2021hift}
Z.~Cao, C.~Fu, J.~Ye, B.~Li, and Y.~Li, ``{HiFT: Hierarchical Feature
  Transformer for Aerial Tracking},'' in \emph{Proceedings of the IEEE/CVF
  International Conference on Computer Vision (ICCV)}, 2021, pp.
  15\,457--15\,466.

\bibitem{bertinetto2016fully}
L.~Bertinetto, J.~Valmadre, J.~F. Henriques, A.~Vedaldi, and P.~H. Torr,
  ``{Fully-Convolutional Siamese Networks for Object Tracking},'' in
  \emph{Proceedings of the European Conference on Computer Vision (ECCV)},
  2016, pp. 850--865.

\bibitem{zhu2018distractor}
Z.~Zhu, Q.~Wang, B.~Li, W.~Wu, J.~Yan, and W.~Hu, ``{Distractor-Aware Siamese
  Networks for Visual Object Tracking},'' in \emph{Proceedings of the European
  Conference on Computer Vision (ECCV)}, 2018, pp. 101--117.

\bibitem{li2019target}
X.~Li, C.~Ma, B.~Wu, Z.~He, and M.-H. Yang, ``{Target-Aware Deep Tracking},''
  in \emph{Proceedings of the IEEE/CVF Conference on Computer Vision and
  Pattern Recognition (CVPR)}, 2019, pp. 1369--1378.

\bibitem{li2017visual}
S.~Li and D.-Y. Yeung, ``{Visual Object Tracking for Unmanned Aerial Vehicles:
  A Benchmark and New Motion Models},'' in \emph{Proceedings of the AAAI
  Conference on Artificial Intelligence (AAAI)}, vol.~31, no.~1, 2017, pp.
  4140--4146.

\bibitem{vaswani2017nips}
A.~Vaswani, N.~Shazeer, N.~Parmar, J.~Uszkoreit, L.~Jones, A.~N. Gomez,
  {\L}.~Kaiser, and I.~Polosukhin, ``{Attention Is All You Need},'' in
  \emph{Proceedings of the Advances in Neural Information Processing Systems
  (NIPS)}, 2017, pp. 6000--6010.

\bibitem{xing2022siamese}
D.~Xing, N.~Evangeliou, A.~Tsoukalas, and A.~Tzes, ``{Siamese Transformer
  Pyramid Networks for Real-Time UAV tracking},'' in \emph{Proceedings of the
  IEEE/CVF Winter Conference on Applications of Computer Vision (WACV)}, 2022,
  pp. 2139--2148.

\bibitem{sun2020transtrack}
P.~Sun, J.~Cao, Y.~Jiang, R.~Zhang, E.~Xie, Z.~Yuan, C.~Wang, and P.~Luo,
  ``{Transtrack: Multiple Object Tracking with Transformer},'' \emph{arXiv
  preprint arXiv:2012.15460}, pp. 1--11, 2020.

\bibitem{wang2021transformer}
N.~Wang, W.~Zhou, J.~Wang, and H.~Li, ``{Transformer Meets Tracker: Exploiting
  Temporal Context for Robust Visual Tracking},'' in \emph{Proceedings of the
  IEEE/CVF Conference on Computer Vision and Pattern Recognition (CVPR)}, 2021,
  pp. 1571--1580.

\bibitem{li2020autotrack}
Y.~Li, C.~Fu, F.~Ding, Z.~Huang, and G.~Lu, ``{AutoTrack: Towards
  High-performance Visual Tracking for UAV with Automatic Spatio-Temporal
  Regularization},'' in \emph{Proceedings of the IEEE/CVF Conference on
  Computer Vision and Pattern Recognition (CVPR)}, 2020, pp. 11\,923--11\,932.

\bibitem{fu2021correlation}
C.~Fu, B.~Li, F.~Ding, F.~Lin, and G.~Lu, ``{Correlation Filters for Unmanned
  Aerial Vehicle-Based Aerial Tracking: a Review and Experimental
  Evaluation},'' \emph{IEEE Geoscience and Remote Sensing Magazine}, vol.~10,
  no.~1, pp. 125--160, 2022.

\bibitem{li2018high}
B.~Li, J.~Yan, W.~Wu, Z.~Zhu, and X.~Hu, ``{High Performance Visual Tracking
  With Siamese Region Proposal Network},'' in \emph{Proceedings of the IEEE
  conference on computer vision and pattern recognition (CVPR)}, 2018, pp.
  8971--8980.

\bibitem{cao2021siamapn++}
Z.~Cao, C.~Fu, J.~Ye, B.~Li, and Y.~Li, ``{SiamAPN++: Siamese Attentional
  Aggregation Network for Real-Time UAV Tracking},'' in \emph{Proceedings of
  the IEEE/RSJ International Conference on Intelligent Robots and Systems
  (IROS)}, 2021, pp. 3086--3092.

\bibitem{chen2021transformer}
X.~Chen, B.~Yan, J.~Zhu, D.~Wang, X.~Yang, and H.~Lu, ``{Transformer
  Tracking},'' in \emph{Proceedings of the IEEE/CVF Conference on Computer
  Vision and Pattern Recognition (CVPR)}, 2021, pp. 8126--8135.

\bibitem{fu2021onboard}
C.~Fu, Z.~Cao, Y.~Li, J.~Ye, and C.~Feng, ``{Onboard Real-Time Aerial Tracking
  With Efficient Siamese Anchor Proposal Network},'' \emph{IEEE Transactions on
  Geoscience and Remote Sensing}, vol.~60, pp. 1--13, 2021.

\bibitem{borji2019salient}
A.~Borji, M.-M. Cheng, Q.~Hou, H.~Jiang, and J.~Li, ``{Salient Object
  Detection: A Survey},'' \emph{Computational Visual Media}, vol.~5, no.~2, pp.
  117--150, 2019.

\bibitem{feng2019dynamic}
W.~Feng, R.~Han, Q.~Guo, J.~Zhu, and S.~Wang, ``{Dynamic Saliency-Aware
  Regularization for Correlation Filter-Based Object Tracking},'' \emph{IEEE
  Transactions on Image Processing}, vol.~28, no.~7, pp. 3232--3245, 2019.

\bibitem{zhou2021saliency}
Z.~Zhou, W.~Pei, X.~Li, H.~Wang, F.~Zheng, and Z.~He, ``{Saliency-Associated
  Object Tracking},'' in \emph{Proceedings of the IEEE/CVF International
  Conference on Computer Vision (ICCV)}, 2021, pp. 9866--9875.

\bibitem{rao2021dynamicvit}
Y.~Rao, W.~Zhao, B.~Liu, J.~Lu, J.~Zhou, and C.-J. Hsieh, ``{DynamicViT:
  Efficient Vision Transformers with Dynamic Token Sparsification},''
  \emph{Advances in Neural Information Processing Systems (NIPS)}, vol.~34, pp.
  13\,937--13\,949, 2021.

\bibitem{tang2022quadtree}
S.~Tang, J.~Zhang, S.~Zhu, and P.~Tan, ``{Quadtree Attention for Vision
  Transformers},'' in \emph{Proceedings of the International Conference on
  Learning Representations (ICLR)}, 2022, pp. 1--16.

\bibitem{jang2017categorical}
E.~Jang, S.~Gu, and B.~Poole, ``Categorical reparameterization with
  gumbel-softmax,'' in \emph{Proceedings of the International Conference on
  Learning Representations (ICLR)}, 2017, pp. 1--12.

\bibitem{liu2021swin}
Z.~Liu, Y.~Lin, Y.~Cao, H.~Hu, Y.~Wei, Z.~Zhang, S.~Lin, and B.~Guo, ``{Swin
  Transformer: Hierarchical Vision Transformer Using Shifted Windows},'' in
  \emph{Proceedings of the IEEE/CVF International Conference on Computer Vision
  (ICCV)}, 2021, pp. 10\,012--10\,022.

\bibitem{mueller2016benchmark}
M.~Mueller, N.~Smith, and B.~Ghanem, ``{A Benchmark and Simulator for UAV
  Tracking},'' in \emph{Proceedings of the European Conference on Computer
  Vision (ECCV)}, 2016, pp. 445--461.

\bibitem{krizhevsky2017imagenet}
A.~Krizhevsky, I.~Sutskever, and G.~E. Hinton, ``{Imagenet Classification with
  Deep Convolutional Neural Networks},'' \emph{Communications of the ACM},
  vol.~60, no.~6, pp. 84--90, 2017.

\bibitem{lin2014microsoft}
T.-Y. Lin, M.~Maire, S.~Belongie, J.~Hays, P.~Perona, D.~Ramanan,
  P.~Doll{\'a}r, and C.~L. Zitnick, ``{Microsoft Coco: Common Objects in
  Context},'' in \emph{Proceedings of the European Conference on Computer
  Vision (ECCV)}, 2014, pp. 740--755.

\bibitem{russakovsky2015imagenet}
O.~Russakovsky, J.~Deng, H.~Su, J.~Krause, S.~Satheesh, S.~Ma, Z.~Huang,
  A.~Karpathy, A.~Khosla, M.~Bernstein \emph{et~al.}, ``{Imagenet Large Scale
  Visual Recognition Challenge},'' \emph{International Journal of Computer
  Vision}, vol. 115, no.~3, pp. 211--252, 2015.

\bibitem{huang2019got}
L.~Huang, X.~Zhao, and K.~Huang, ``{Got-10k: A Large High-Diversity Benchmark
  for Generic Object Tracking in the Wild},'' \emph{IEEE Transactions on
  Pattern Analysis and Machine Intelligence}, vol.~43, no.~5, pp. 1562--1577,
  2019.

\bibitem{real2017youtube}
E.~Real, J.~Shlens, S.~Mazzocchi, X.~Pan, and V.~Vanhoucke,
  ``{YouTube-BoundingBoxes: A Large High-Precision Human-Annotated Data Set for
  Object Detection in Video},'' in \emph{proceedings of the IEEE Conference on
  Computer Vision and Pattern Recognition (CVPR)}, 2017, pp. 5296--5305.

\bibitem{muller2018trackingnet}
M.~Muller, A.~Bibi, S.~Giancola, S.~Alsubaihi, and B.~Ghanem, ``{Trackingnet: A
  Large-Scale Dataset and Benchmark for Object Tracking in the Wild},'' in
  \emph{Proceedings of the European Conference on Computer Vision (ECCV)},
  2018, pp. 300--317.

\bibitem{zhang2019deeper}
Z.~Zhang and H.~Peng, ``{Deeper and Wider Siamese Networks for Real-Time Visual
  Tracking},'' in \emph{Proceedings of the IEEE/CVF Conference on Computer
  Vision and Pattern Recognition (CVPR)}, 2019, pp. 4591--4600.

\bibitem{zhang2020ocean}
Z.~Zhang, H.~Peng, J.~Fu, B.~Li, and W.~Hu, ``{Ocean: Object-aware Anchor-free
  Tracking},'' in \emph{Proceedings of the European Conference on Computer
  Vision (ECCV)}, 2020, pp. 771--787.

\bibitem{sosnovik2021scale}
I.~Sosnovik, A.~Moskalev, and A.~W. Smeulders, ``{Scale Equivariance Improves
  Siamese Tracking},'' in \emph{Proceedings of the IEEE/CVF Winter Conference
  on Applications of Computer Vision (WACV)}, 2021, pp. 2765--2774.

\bibitem{huang2019learning}
Z.~Huang, C.~Fu, Y.~Li, F.~Lin, and P.~Lu, ``{Learning Aberrance Repressed
  Correlation Filters for Real-Time UAV Tracking},'' in \emph{Proceedings of
  the IEEE/CVF International Conference on Computer Vision (ICCV)}, 2019, pp.
  2891--2900.

\bibitem{wang2019unsupervised}
N.~Wang, Y.~Song, C.~Ma, W.~Zhou, W.~Liu, and H.~Li, ``{Unsupervised Deep
  Tracking},'' in \emph{Proceedings of the IEEE/CVF Conference on Computer
  Vision and Pattern Recognition (CVPR)}, 2019, pp. 1308--1317.

\bibitem{guo2017learning}
Q.~Guo, W.~Feng, C.~Zhou, R.~Huang, L.~Wan, and S.~Wang, ``{Learning Dynamic
  Siamese Network for Visual Object Tracking},'' in \emph{Proceedings of the
  IEEE International Conference on Computer Vision (ICCV)}, 2017, pp.
  1763--1771.

\bibitem{zhang2019learning}
L.~Zhang, A.~Gonzalez-Garcia, J.~v.~d. Weijer, M.~Danelljan, and F.~S. Khan,
  ``{Learning the Model Update for Siamese Trackers},'' in \emph{Proceedings of
  the IEEE/CVF International Conference on Computer Vision (ICCV)}, 2019, pp.
  4010--4019.

\end{thebibliography}


\end{document}